\documentclass[journal]{IEEEtran}
\usepackage{amsmath}
\usepackage{color}
\usepackage[ruled,linesnumbered, vlined]{algorithm2e}
\makeatletter
\newcommand{\removelatexerror}{\let\@latex@error\@gobble}
\makeatother

\usepackage{flushend}
\usepackage{array}
\usepackage{graphicx}
\usepackage{amsmath,amsfonts} 
\usepackage{multirow} 
\usepackage{booktabs} 

\ifCLASSOPTIONcompsoc
 \usepackage[caption=false,font=normalsize,labelfont=sf,textfont=sf]{subfig}
\else
 \usepackage[caption=false,font=footnotesize]{subfig}
\fi

\usepackage{stfloats}
\usepackage{url}
\begin{document}
\title{DynG2G: An Efficient Stochastic Graph Embedding Method for Temporal Graphs}

\author{Mengjia Xu,
        Apoorva Vikram Singh, 
        and~George Em Karniadakis%
\thanks{This work was supported by Swiss Robotics and by Brown University's cost sharing to ARO MURI grant W911NF-15-1-0562; High Performance Computing resources were provided by the Center for Computation and Visualization at Brown University.} 
\thanks{M. Xu and G.E. Karniadakis are with the Division of Applied Mathematics, Brown University, Providence, RI 02912, USA; M. Xu is also with McGovern Institute for Brain Research, Massachusetts Institute of Technology, Cambridge, MA 02139, USA (e-mails: george\_karanidaskis@brown.edu and mengjia@mit.edu).}
\thanks{A.V. Singh is with the Department of Electrical Engineering, National Institute of Technology, Silchar, Assam 788010, India (e-mail: singhapoorva388@gmail.com).
}
}
\maketitle

\begin{abstract}
Dynamic graph embedding has gained great attention recently due to its capability of learning low-dimensional and meaningful graph representations for complex temporal graphs with high accuracy. However, recent advances mostly focus on learning node embeddings as \textit{deterministic ``vectors''} for {\em static}  graphs, hence disregarding the key graph temporal dynamics and the evolving uncertainties associated with node embedding in the latent space. In this work, we propose an efficient stochastic dynamic graph embedding method (DynG2G) that applies an inductive feed-forward encoder trained with node triplet energy-based ranking loss. Every node per timestamp is encoded as a time-dependent \textit{probabilistic multivariate Gaussian distribution} in the latent space, hence
we are able to quantify the node embedding uncertainty on-the-fly. We have considered eight different benchmarks that represent diversity in size (from 96 nodes to 87,626 and from 13,398 edges to 4,870,863) as well as diversity in dynamics, from slowly changing temporal evolution to rapidly varying multi-rate dynamics. 
We demonstrate through extensive experiments based on these eight dynamic graph benchmarks that DynG2G achieves new state-of-the-art performance in capturing the underlying temporal node embeddings. We also demonstrate that DynG2G can simultaneously predict the evolving node embedding uncertainty, which plays a crucial role in quantifying the {\em intrinsic dimensionality} of the dynamical system over time. In particular, we obtain a ``universal" relation of the optimal embedding dimension, $L_o$, versus the effective dimensionality of uncertainty, $D_u$, and we infer that $L_o=D_u$ for all cases; see Table III. This, in turn, implies that the uncertainty quantification approach we employ in the DynG2G algorithm correctly captures the {\em intrinsic dimensionality} of the dynamics of such evolving graphs despite the diverse nature and composition of the graphs at each timestamp. In addition, this $L_0 - D_u$ correlation provides a clear path to selecting adaptively the optimum embedding size at each timestamp by setting $L \ge D_u$.
\end{abstract}


\IEEEpeerreviewmaketitle

\section{Introduction}
\IEEEPARstart{N}{umerous} real-world systems encompassing complex behavior and interaction among entities can be naturally modeled as ``graphs''~\cite{wu2020comprehensive}. Recently, deep neural network-based graph embedding techniques have proven very promising and effective in learning highly informative graph representations for various graph analytics problems in diverse fields~\cite{xu2020understanding}, e.g., social network analysis~\cite{wang2019user}, protein-protein interaction prediction~\cite{ashoor2020graph}, functional brain network analysis~\cite{xu2020new, xu2021graph}, electronic health record data analysis~\cite{choi2020learning}, molecular property prediction in chemistry~\cite{rong2020self}, tumor micro-environment staging for prognosis prediction of the gastric cancer~\cite{wang2021cell}, high energy physics data analysis~\cite{shlomi2020graph}, etc. Specifically, by leveraging neural networks, such as SkipGram~\cite{mikolov2013efficient}, autoencoder~\cite{wang2016structural} and graph convolutional networks~\cite{kipf2016semi}, high-dimensional graphs can be effectively transformed into low-dimensional latent space as continuous vectors (i.e., node embeddings), while the original graph topological properties are maximally preserved. Moreover, the node embedding distances in the latent space can be utilized to approximate the node similarity in the original irregular space. To this end, the obtained graph embeddings can benefit a variety of downstream graph analyses tasks, e.g., node classification, link prediction, anomaly detection, recommender systems, etc. Nevertheless, many of the aforementioned high-dimensional graphs in different applications exhibit heterogeneous {\em evolving} topologies and {\em varying} features at the {\em temporal scale}. Hence, {\em temporal graph embedding methods} have surged to become a hot topic due to their key advantages in capturing graph representations at both spatial {\em and} temporal scales. However, how to efficiently embed the dynamic graphs still remains quite challenging requiring to address the following three main questions:

\begin{itemize}
    \item How to stabilize the graph embedding training caused by heterogeneous graph dynamics in both topology and feature occurred in the temporal graphs;
    \item How to estimate the uncertainty for node embeddings so that it provides robustness for fine-grained quantification of node properties in the latent space;
    \item How to reduce computational cost and enhance scalability for graphs.
\end{itemize}

To tackle the aforementioned issues, we present a new stochastic graph embedding method (DynG2G) for learning probabilistic temporal graph representations and enabling node uncertainty quantification. The proposed DynG2G model follows four main steps: 1) Sampling node triplets based on the k-hop neighborhood incorporating edge weights; 2) Developing an adaptive feed-forward encoder to learn graph embeddings for temporal graphs with varying node numbers in different graph snapshots~\footnote{The number of the hidden units in our DynG2G encoder can be adaptively assigned based on the changing node numbers in different graph snapshots}; 3) Initializing the deep encoder for the current timestamp with the learned hyper-parameters from the previous graph snapshots; 4) Training and optimizing the neural network with a \textit{triplet energy-based square-exponential contrastive} loss. 

To demonstrate the effectiveness of DynG2G, we test it for temporal link prediction on eight different benchmarks, varying in size of nodes and edges but also varying in the intrinsic dynamics from very stable graphs to rapidly changing graphs as a function of time.
Comparisons with baselines, when available, demonstrate that DynG2G can achieve favorable predictive performance and high efficiency in temporal link prediction on different benchmarks by using the obtained dynamic probabilistic node embeddings (i.e., multivariate Gaussian distribution in terms of mean and variance vectors). Additionally, DynG2G provides an important capability of uncertainty quantification for the node embedding, which allows us to produce more reliable and quantitative graph representations in the latent space for various downstream tasks. Moreover, it reveals the (possibly evolving) effective dimensionality of the complex dynamical system that we represent as we demonstrate in our experiments for all eight benchmarks. This, in turn, relates to the optimum embedding size, which for problems with highly transient dynamics may be changing over time.

The main contributions of our paper are as follows.
\begin{itemize}
    \item We developed a \textit{stochastic} graph embedding model (DynG2G) that projects temporal graphs from irregular domain to low-dimensional ``function'' space, which allows to capture node embedding uncertainty;
    \item   DynG2G produces more informative node embeddings that achieve superior performance in temporal link prediction task, for all benchmarks and especially for the highly dynamic benchmarks, e.g., UCI and Bit-OTC datasets, see details in Section \ref{p_linkpredict}.
    \item We discovered a ``universal relation", a correlation that connects the optimum embedding dimension with the effective dimension of the dynamic system, hence identifying a clear way to select the optimum embedding dimension using uncertainty quantification, see results in Table.~\ref{tab:DvsL}. 
\end{itemize}

We have posted our DynG2G code \footnote{\url{https://github.com/GraceXu182/DynG2G}} on the GitHub so that the interested reader can reproduce the results.

\section{Related work}
Dynamic graph embedding is a key and prerequisite step for temporal graph machine learning. It has attracted an increasing interest in dynamic graph embedding studies recently due to its superiority in learning latent graph representations for many different tasks. There are four main types of approaches:

\subsection{Matrix factorization-based models} 
For this approach, \cite{zhu2018high} proposed the DHPE model that mainly applies the eigen-decomposition to construct the high-order proximity matrix of the network, and then dynamically updates the node embedding of the next snapshot via the matrix perturbation approach~\cite{li2015analysis}. However, matrix factorization-based models usually suffer from high computational complexity for large scale temporal graphs; moreover, the incremental matrix decomposition procedure is prone to error accumulation. \cite{zhang2018timers} proposed another more advanced TIMERS model, which effectively reduces the error accumulation problem by using an error bound-based constraint during node embedding updating over time.

\subsection{SkipGram-based models} 
The SkipGram model (aka word2vec) was originally developed as a random walk-based language embedding approach, and it has been very successfully applied in diverse static graph embedding models, e.g., node2vec~\cite{grover2016node2vec}, deepwalk~\cite{perozzi2014deepwalk}. The majority of dynamic graph embedding approaches focus on leveraging the conventional static graph embedding models incorporating temporal information~\cite{singer2019node, mahdavi2018dynnode2vec, du2018dynamic, yu2018netwalk}. In dynnode2vec~\cite{mahdavi2018dynnode2vec} the authors adopted the learned parameters in the previous snapshot for network initialization in the next snapshot. They trained the new evolving random walks and updated the node embedding for the evolving nodes at each time step. Hence,  dynnode2vec can take advantage of the previous learned mappings to incorporate the important temporal dependencies features to predict the links in the next time step $t$; dynnode2vec~\cite{mahdavi2018dynnode2vec} outperforms the conventional static graph embedding model (i.e., node2vec~\cite{grover2016node2vec}) for anomaly detection. In addition to learning embeddings for discrete graph snapshots using conventional static graph embedding models, the CTDNE model~\cite{nguyen2018dynamic} employs a continuous-time dynamic graph embedding approach, which enables learning of the time-preserving graph embeddings using the SkipGram model with valid temporal walks ($W_T$) dynamic graph streams (a scale of seconds or milliseconds). 

\subsection{Autoencoder-based models}
The unsupervised learning-based deep autoencoder has shown great effectiveness in learning static graph embeddings and is particularly applicable to graph reconstruction problems. \cite{goyal2018dyngem} proposed a ``semi-supervised'' dynamically expansion autoencoder model (DynGEM) for discrete-time growing graph embedding. To achieve the optimal graph embeddings over time, it optimizes the parameters of deep autoencoder for each snapshot through minimizing a weighted combination loss built by the embedding loss and the graph reconstruction loss, which can help to preserve both local and global graph structure properties. However, DynGEM only considers temporal patterns spanning in two consecutive time snapshots. Moreover, it assumes that the graph dynamic changes are smooth and use regularization terms to disallow rapid changes. To address these problems, \cite{goyal2020dyngraph2vec} further proposed a dyngraph2vec model including three variants (dyngraphAE, dyngraphRNN and dyngraphAERNN) for learning dynamic graph series embeddings over long sequences of previous snapshots. The main goal of dyngraph2vec~\cite{goyal2020dyngraph2vec} is to predict the graph embedding for the future snapshot incrementally using the past snapshots. A recurrent neural network (RNN) layer was added to learn the temporal dynamics across different graph snapshots. Hence, dyngraph2vec~\cite{goyal2020dyngraph2vec} adopted a look back parameter ($l = 1,2,3$) to specify the number of look back time steps for predicting the next snapshot. However, the model is large and computationally expensive. 

\subsection{GNN-based models}
With the great success of graph neural networks (GNN) in static graph representation learning, a few recent studies proposed time-dependent GNN models integrating GNN and RNN (or LSTM) for learning temporal graph sequence representations in different graph snapshots, e.g., TDGNN~\cite{qu2020continuous}, EvolveGCN~\cite{pareja2020evolvegcn}. More recently, \cite{sankar2020dysat, rossi2020temporal} adopted the self-attention mechanism to jointly encode the structural and temporal dynamics for temporal graph sequences. The self-attention mechanism helps to effectively capture long-range dependencies and draw most relevant context from all past graph snapshots to adaptively assign interpretable weights for previous time steps. Moreover, there are also a few works modeling the dynamic graphs as temporal point processes in conjunction with the attention mechanism, e.g., DynamicTriad~\cite{zhou2018dynamic}, HTNE~\cite{zuo2018embedding}, DyRep\cite{trivedi2019dyrep}. More dynamic graph embedding literature can be found in the surveys~\cite{goyal2020graph,ji2021survey, xu2020understanding,xie2020survey}.

Nevertheless, the majority of existing dynamic graph embedding models are ``deterministic'', hence disregarding the important ``uncertainty information'' for the graph embeddings in the latent space. As we will demonstrate herein, the uncertainty quantification is ultimately related to the embedding dimension of the system, with the effective dimension of the uncertainty being a lower bound for the optimum embedding dimension. This is a key finding of our work and is verified for all eight diverse benchmarks we consider in our current study. 

\section{Methodology}
\subsection{Notation} 
\begin{figure*}[!t]
\centering
\includegraphics[width=.85\textwidth]{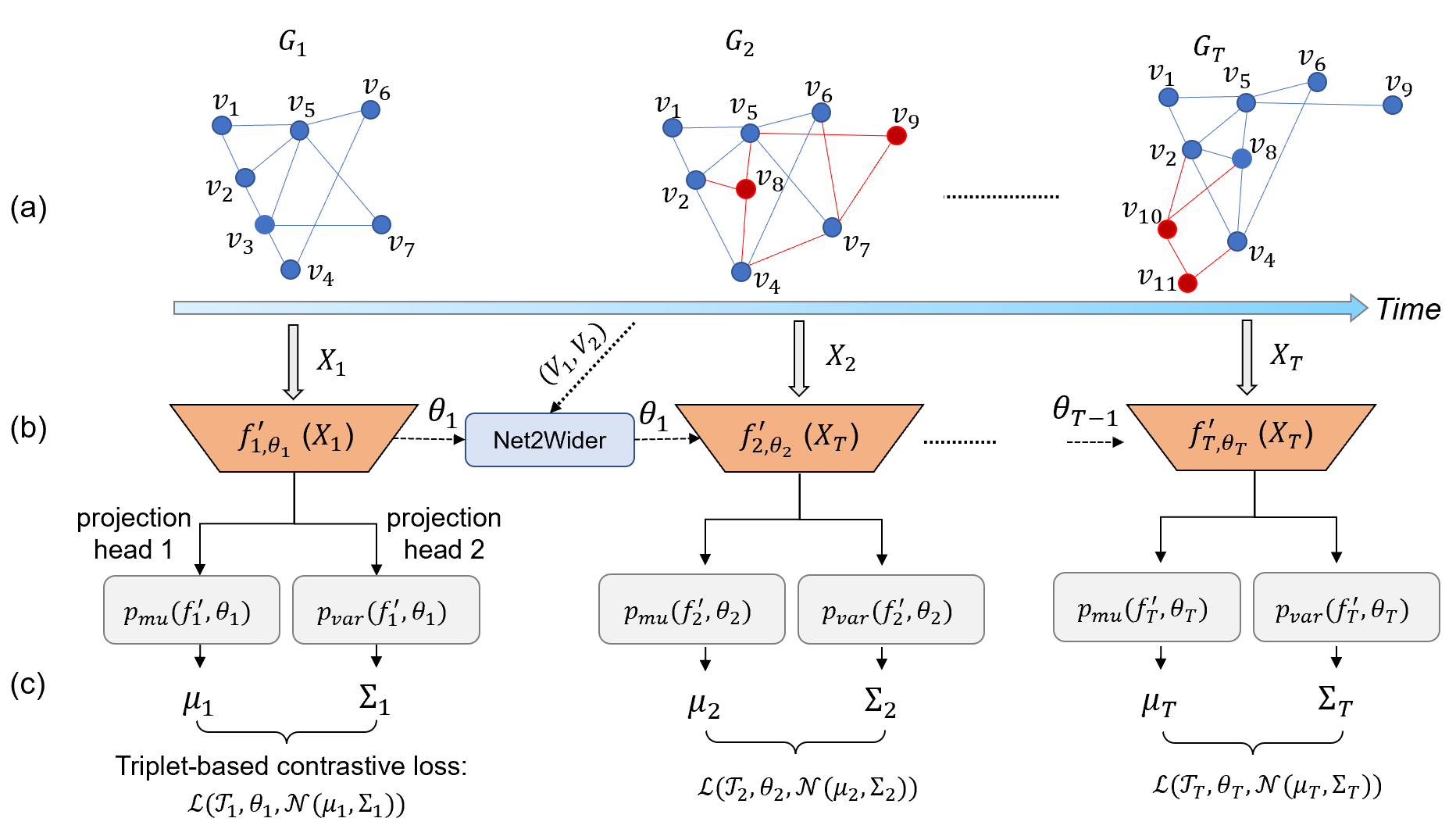}
\caption{Schematic diagram of our proposed DynG2G framework for stochastic temporal graph embedding. The dynamic graph $G$ is composed of $T$ graph snapshots. To learn the latent graph representations for the $\mathcal{G} = \{G_1, G_2, ..., G_T\}$ over $T$ timestamps (vertices and links in red are the new added ones), we first generate for each node a set of node triplets ($\mathcal{T}_{1}$) based on the $k$-hop node neighbor sets computed from the adjacency matrix $A_1$ of $G_1$. The attribute matrix $X_1$ (``one-hot encoding'' of the nodes if a non-attributed graph) at the first timestamp is input to an encoder $f_{1}^{'}(\cdot)$ with parameters $\theta_1$ and two separate projection layers ($p_{mu}$ and $p_{var}$), yielding the snapshot-level graph node embeddings in terms of mean vector $\mu_1^i \in \mathbb{R}^{L}$ and diagonal covariance matrix $\Sigma_1^i \in \mathbb{R}^{L\times L}$ for each node $v_i$ ($L$ is the embedding size). For the next snapshot training, we employ an extension of the Net2WiderNet approach~\cite{chen2015net2net} to adaptively expand the network hidden layer size based on the number of changing nodes in the next snapshot. To effectively capture the temporal graph dynamics across different graph snapshots, we train the encoder $f_{2}(\cdot)^{'}$ in the second timestamp with the hyper-parameters ($\theta_2$) transferred from the pre-trained model for the first graph snapshot. We trained the model with a time-dependent node triplet-based contrastive loss $\mathcal{L} (\mathcal{T}_t, \theta_t, \mathcal{N}(\mu_t, \Sigma_t))$.} 
\label{fig:workflow} 
\end{figure*}

We model a dynamic graph as a series of $T$ graph snapshots $\mathcal{G} = \{G_1, G_2, ..., G_T\}$, see an example in Fig.~\ref{fig:workflow}(a). The graph snapshot $G_t$ at timestamp $t$ consists of a vertex set $V_t = \{v_1, v_2,..., v_{|\mathcal{V}_t|}\}$ of $|\mathcal{V}_t|$ nodes and an edge set $E_t = \{e_{i,j}| i,j \in |\mathcal{V}_t|\}$, where each edge $e_{i,j}$ in the graph connects two vertices $v_i$ and $v_j$. $A_t$ represents the corresponding adjacent matrix of graph snapshot $G_t$, which can be either weighted or unweighted, directed or undirected. The node features are denoted by $X_t \in \mathbb{R}^{\mathcal{V}_{t} \times D}$, where $D$ is the dimensionality of the node attributes. The main goal of stochastic temporal graph embedding is to learn time-dependent graph mappings ($\mathcal{F} = \{f_1, f_2, ..., f_T\}$) for different graph snapshots, such that each graph node can be represented as a sequence of lower-dimensional multivariate Gaussian distributions.

\subsection{Problem Formulation}
Herein, motivated by the previous work ``Graph2Gauss (G2G)''~\cite{bojchevski2017deep} that only focuses on learning graph Gaussian embeddings for static graphs, we propose an efficient and scalable graph Gaussian embedding framework (``DynG2G'') for more complex \textit{time-evolving} graphs. The main goal of our work is to learn temporal graph embeddings as low-dimensional statistical distributions by incorporating the important graph evolutionary information from the previous graph sequences. In particular, the proposed DynG2G is capable of projecting a temporal graph from a non-Euclidean space to low-dimensional time-dependent ``function'' space (unlike the conventional ``deterministic vector'' space introduced in most existing studies~\cite{goyal2018dyngem, goyal2020dyngraph2vec, pareja2020evolvegcn, zhou2021grassmann}) via a series of stochastic mappings ($\mathcal{F}^{'} = \{f_{1}^{'}, f_{2}^{'}, ..., f_{T}^{'}\}$) as shown in Fig.~\ref{fig:workflow}{(b)}. Each graph mapping $f_{t}^{'}$ transforms the graph snapshot ($G_{t}$) at timestamp $t$ to a lower-dimensional multivariate Gaussian distributions $H_t = \{h_{t}^i| i \in |\mathcal{V}_t|\}$, where each node can be represented as a joint normal distribution $h_t^{i} = \mathcal{N}(\mu_i, \Sigma_i)$ in terms of a mean vector $\mu_i \in \mathbb{R}^{L}$ and a covariance matrix $\Sigma_i \in \mathbb{R}^{L\times L}$ ($L$ is the embedding size). Finally, both spatial graph structure properties and temporal graph dynamics can be maximally preserved in the latent space. 

\subsection{DynG2G: A stochastic temporal graph embedding framework with uncertainty quantification}
The main workflow of the DynG2G is shown in Fig.~\ref{fig:workflow}, involving four key phases: 1) Generation of node triplet based on weighted k-hop sampling for every graph snapshot as shown in Fig.~\ref{fig:workflow}(a); 2) Building a snapshot-level stochastic graph embedding model based on a simplified nonlinear feed-forward neural network encoder ($f^{'}$) with one `` hidden layer'' followed by two separate projection layers ($p_{mu}$ and $p_{var}$) for outputting the final lower-dimensional multivariate Gaussian distributions, i.e., mean and variance vectors, respectively; 3) Training recurrently the DynG2G model for the graph snapshot $G_{t+1}$ at timestamp $t+1$ by transferring the pre-trained parameters from $G_t$ at the previous timestamp $t$; 4) Optimization of the model with the node triplet-based contrastive ranking loss. 

\subsubsection{Generation of node triplet based on K-hop neighboring sampling approach} 
In order to preserve graph topological properties in the latent space, we first sample the node neighbors based on different $k \, (k = 1, 2, 3, ...K)$ hops, e.g., $N_{ik}$ denotes the k-hop node neighbors of node $v_i$. The $K$ node hops (``node context'') sampled for each node in the graph snapshot are then used to generate the corresponding node triplet set ($\mathcal{T}_t = \{(v_{i}, v_{i}^{+}, v_{i}^{-})|v_{i} \in V_{t}\}$), where $v_{i}$ is the anchor node, $v_{i}^{+}$ and $v_{i}^{-}$ are its positive nodes and negative nodes, respectively. The shortest path ($sp(.)$) between positive node pair ($v_{i}$ and $v_{i}^{+}$) and the shortest path between the negative node pairs ($v_{i}$ and $v_{i}^{-}$) follow the constraint: $sp(v_{i}, v_{i}^{+}) < sp(v_i, v_{i}^{-})$. In other words, positive nodes ($v_{i}^{+}$) are much closer (i.e., within smaller hops) to the anchor node $v_i$ than the negative nodes $v_{i}^{-}$. In order to achieve multi-scale graph structure property preservation in the latent space, we ultimately rank the KL-divergence dissimilarity between the latent node Gaussian embeddings over the sampled different $K$-hops for each node. The detailed node pair energy ranking formulas is shown in Eq.~\ref{eq:sample_triplets}:
\begin{equation}
\begin{split}
&E(P_i, P_{k_1}) < E(P_i, P_{k_2}) < E(P_i, P_{k_3})...< E(P_i, P_{k_K}),\\
&\forall k_1 \in N_{i1}, k_2 \in N_{i2}, ..., k_K \in N_{iK}.
\end{split}
\label{eq:sample_triplets}   
\end{equation}

\subsubsection{Stochastic temporal contrastive graph embedding learning}
In comparison with the previous works that transform temporal graphs as deterministic ``low-dimensional'' vectors, in our work, we propose a \textit{stochastic} temporal contrastive graph embedding learning method, which can transform each node of a graph snapshot into a density-based function space in terms of a \textit{multivariate Gaussian distribution} via a nonlinear feed-forward neural network ($f_{t}^{'}$ with one hidden layer of 512 units) and two separate projection heads ($p_{mu}$ and $p_{var}$), which share the same parameters ($\theta_t$) as the encoder $f_{t}^{'}$) to firstly learn the hidden representation for each graph node and then output the final stochastic graph node Gaussian embeddings ($\mathcal{N}_{t}(\mu_{t}, \Sigma_{t})$) at timestamp $t$ in terms of the mean vector ($\mu_t \in \mathbb{R}^{L}$) and the covariance matrix ($\Sigma_t \in \mathbb{R}^{L\times L}$) (the diagonal elements are the ``variances''), $L$ is the node embedding size. More details about the snapshot-level encoding procedure can be seen from the first column of Fig.~\ref{fig:workflow}.

Subsequently, aiming to make our DynG2G model encode temporal graph patterns from the previous graph snapshot ($G_{t-1}$) to the next one $G_{t}$, we first input the weights to the Net2WiderNet model~\cite{chen2015net2net} and widen the encoder input layer size according to the increasing number of nodes in the $G_{t}$. 
Following \cite{chen2015net2net}, we assume that graph snapshots $G_{t-1}$ and $G_{t}$ have node numbers $V_{t-1}$ and $V_{t}$, where $V_{t} > V_{t-1}$. After training the G2G encoder $f_{t-1}^{'}(\cdot)$ for $G_{t-1}$, we need to widen the input layer size of the encoder $f_{t}^{'}(\cdot)$ for the next timestamp based on the increased number of nodes ($V_{t}-V_{t-1}$); then $\theta_{t-1}$ is replaced
by $\theta_{t}$. If $G_{t-1}$ has $V_{t-1}$ inputs and 512 outputs, and timestamp $t$
has $V_{t}$ inputs, then $\theta_{t-1} \in R^{V_{t-1}\times 512}$ and $\theta_{t} \in R^{V_{t}\times 512}$. Net2WiderNet enables the encoder to increase the layer size that has $V_t$ inputs, with $V_t > V_{t-1}$. By defining a random mapping function $g: \{1, 2, · · · , V_{t}\} \mapsto \{1, 2, ... , V_{t-1}\}$, that satisfies
$g(i) = j$ if $j \le  V_{t-1}$ else $g(i) =$ random sample from $\{1,2,...,  V_{t-1}\}$, then we define a new weight matrix $U_{t-1}$ and $U_{t}$ representing the weights for these new layers. Hence, the new weights in the next timestamp are obtained by
\begin{equation}
    U_{t-1}^{j} = \theta_{t-1}^{g(j)},\,\,  U_{t}^{j} = \frac{1}{|\{x|g(x) = g(j)\}|}\theta_{t}^{g(j)},
\end{equation}
where the first $ V_{t-1}$ columns of $\theta^{(t-1)}$ are copied directly into $U_{t-1}$. Columns $ V_{t-1}+1$ through $ V_{t}$ of $U_{t-1}$
are created by choosing a random mapping as defined in $g$ by replacement. Specifically, each column of $\theta_{t-1}$
may be copied multiple times. Regarding the weights in $U_{t}$, we could account
for the multiplicity by dividing the weight by a factor given by $\frac{1}{|\{x|g(x) = g(j)\}|}$.
\vspace{0.05in}

Therefore, the DynG2G architecture transfers the pre-trained encoder's parameters in the previous timestamp for the initialization of the encoder's parameters at the current timestamp, which allows for more efficient and stable training and effectively captures the inter-snapshot evolving patterns. Finally, with the obtained graph node Gaussian distribution embeddings in the latent space, we can build the objective function as a form of ``triplet-based contrastive loss'' (see Eq.~\ref{eq:loss}). In the latent representation space, we use the KL-divergence measure to quantify the difference between two nodes' Gaussian distributions. The loss function contains two main terms: one is the \textit{positive pairs' energy} term measuring the ``KL-divergence'' dissimilarity between the positive node pairs sampled from each graph snapshot; the other one is the negative pairs' energy term for measuring the ``KL-divergence'' dissimilarity between the negative node pairs. The detailed formulas are shown in Eq.~\ref{eq:kl_divergence}.
Note that a distinct property of our DynG2G method is that the obtained node embedding variance allows us to quantify important uncertainty information for temporal graph node embeddings in the latent space. Hence, our proposed DynG2G framework enables us to quantitatively analyze the time-dependent dynamics as well as the transition complexity of the dynamic system over time.  
 
 \subsubsection{Model optimization with node triplet-based contrastive loss}
In order to optimize the DynG2G model and achieve the optimal stochastic graph embeddings for dynamic graph $\mathcal{G}$, we implement the DynG2G model in PyTorch and apply the Adam optimizer to minimize the time-dependent ``node triplet-based'' contrastive loss, such that the positive node pairs' energy (i.e., ``KL-divergence'') are minimized, and the negative node pairs' embedding energy can be maximized. Hence, the positive node pairs' embeddings are more similar, while the negative pairs' embeddings are more dissimilar in the latent space.

\begin{equation}
\label{eq:loss}
    \mathcal{L}_{t} = \sum_{(v_i, v_{i}^{+}, v_{i}^{-}) \in \mathcal{T}_t}{[{\mathbb{E}_{(v_i, v_{i}^{+})}^2+e^{-\mathbb{E}_{(v_i, v_{i}^{-})}}}]} ,
\end{equation}
where $\mathbb{E}_{(v_i, v_{i}^{+})}$ and $\mathbb{E}_{(v_i, v_{i}^{-})}$ refer to the Kullback–Leibler (KL) divergence (see Eq.~\ref{eq:kl_divergence}) between the multivariate Gaussian embeddings of positive node pairs and negative node pairs in the node triplet set ($\mathcal{T}_t$), respectively. $\mathcal{T}_t$ is generated by sampling each node's neighbors by k-hop neighborhood introduced in the above step 1)~\cite{bojchevski2017deep}.
\begin{multline}
\label{eq:kl_divergence}
D_{KL}(\mathcal{N}_i(\mu^i, \Sigma^i)||\mathcal{N}_j(\mu^j, \Sigma^j)) = \frac{1}{2}[tr({\Sigma^j}^{-1}\Sigma^i) +\\ (\mu^j-\mu^i)^{T}{\Sigma^j}^{-1}(\mu^j-\mu^i)-L+log\frac{|\Sigma^j|}{|\Sigma^i|}]
\end{multline}
where $tr(\cdot)$ denotes the trace of a matrix. $L$ corresponds to the dimensionality of the node's Gaussian embeddings in the latent space.

The details of the DynG2G algorithm are presented in Algorithm~\ref{algorithm1}. 
\begin{algorithm}[ht]            
\SetAlgoLined
\caption{DynG2G}
\label{algorithm1}
\KwIn{Weights $\theta_{t-1}$ of the network at previous timestamp, Attribute matrix $X_t$, Number of vertices of current timestamp $\mathcal{V}_t$ and previous timestamp $\mathcal{V}_{t-1}$, node triplet set $\mathcal{T}_t$.}
\KwOut{Time-dependent probabilistic embeddings $H_{t}$ = $\{\mathcal{N}(\mu_{t}^{i}, \Sigma_{t}^{i})|i\in \mathbb{R}^{\mathcal{V}_{t}}\}$, where $\mu_{t}^i \in \mathbb{R}^{\mathcal{V}_{t} \times L}$, $\Sigma_{t}^i \in \mathbb{R}^{\mathcal{V}_{t} \times L \times L}.$}
\For{$t \leftarrow 1:T$}{
    \For{each epoch}{
        Create a G2G Encoder model $f_{t}^{'}(\cdot)$ with initialized weights $\theta_{t}$\\
        \eIf{t = 1}{
        Initialize $\theta_{t}$ with random weights\\
        }{
        $\theta_{t} = \theta_{t-1}$
        }
        \If{$\mathcal{V}_{t} > \mathcal{V}_{t-1}$}{
        $f_{t}^{'}(\cdot) \leftarrow Net2Wider(f_{t-1}^{'}(\cdot))$ \\
        }
        $\mu_{t} \leftarrow p_{mu}(f_{t}^{'}, \theta_t)$ \\
        $\Sigma_{t} \leftarrow p_{var}(f_{t}^{'}, \theta_t)$ \\
        Loss = $\mathcal{L}(\mathcal{T}_{t}, \theta_t, \mathcal{N}(\mu_{t},\Sigma_{t}))$ \\
        Backpropogate the loss\\
    }
}
\textbf{Return} $H_t$
\end{algorithm}

We first train the encoder using the graph at the first timestamp $G_{1}$ with random initialization of model parameters $\theta_{1}$. The subsequent timestamps are trained using the encoder with the condition that its initialization will be done using the parameters from the previous timestamp encoder $\theta_{t-1}$. In order to handle the growing graphs, the encoder is ``widened” if the number of nodes increases in the graphs of subsequent timestamps. The ``widening” of encoder allows us to increase the number of neuron units in the input layer of the encoder while preserving the weights of the network from the earlier timestamps of the training. We handle the widening of the encoder by adopting and expanding the Net2WiderNet algorithm \cite{chen2015net2net}. Once the network is trained for the $(t-1)$th timestamp, the network weights are firstly input to Net2WiderNet that widens the hidden layer size of the encoder on the basis of the number of ``new nodes'' (in red in Fig.~\ref{fig:workflow}(a)) added to graph snapshot. The output from the Net2WiderNet gives us the network weights to initialize the encoder at the next timestamp $t$. To this end, using the network parameters from the previous timestamp results in transferring temporal information from $f_{t-1}^{'}$ to $f_{t}^{'}$ effectively and achieves faster convergence of the training for subsequent timestamps. 

\section{Experiments and analysis}
\label{sec:experiments}
\subsection{Dataset description}
\label{sec:dataset}
\begin{figure*}[ht]
    \centering
    \includegraphics[width=0.8\textwidth]{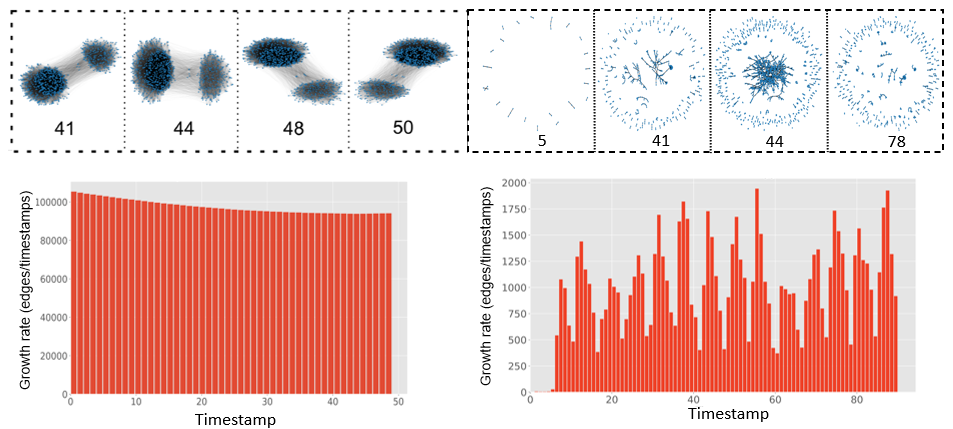}
    \caption{Visualization of the evolving dynamics of two temporal graphs for two of our eight benchmarks: synthetic SBM dataset (left) and Digg (right). Top row: snapshots of the two datasets at different timestamps (indicated by the number in each image). Bottom row: Growth rate of edges as a function of time. SBM (small set) exhibits stable dynamics whereas Digg (large set) exhibits rapidly changing dynamics. Similar visualizations of the dynamics of the other six benchmarks are shown in Appendix Fig. S1.}
    \label{fig:dataset}
\end{figure*}

We evaluated our DynG2G model on eight different benchmarks with different temporal dynamics. The specific graph dataset statistics are shown in Table~\ref{tab:benchmark_table}. Fig.~\ref{fig:dataset} shows two temporal graph examples for the typical snapshots in SBM and Digg datasets. 
\begin{table}[ht]
\centering
\caption{Experiment dataset description.}
\label{tab:benchmark_table}
\begin{tabular}{lllll}
    \toprule
     Dataset & \#Nodes & \#Edges & \#Timestamps   & \#Train/Val/Test \\
     \midrule
     AS & 6,474 & 13,895 & 100 & 70/10/20\\
     SBM & 1,000 & 4,870,863 & 50 & 35/5/10\\
     Bitcoin-OTC & 5,881 & 35,588 & 137 & 95/14/28\\
     UCI & 1,899 & 59,835 & 88 & 62/9/17\\
     Slashdot & 50,824 & 42,968 & 12 & 8/2/2\\
     Facebook & 46,873 & 857,815 & 30 & 21/3/6\\
     Reality Mining & 96 & 1,086,403 & 90 & 54/9/18\\
     Digg & 87,626 & 30,398 & 90 & 54/9/18\\
     \bottomrule
\end{tabular}
\end{table}

\textbf{Autonomous Systems (AS) dataset\footnote{\url{https://snap.stanford.edu/data/as-allstats.html}}}: It consists of a communication network of who-talks-to-whom from the BGP (Border Gateway Protocol) logs. The dataset can be used for predicting message exchange in the future. The AS dataset used in our experiment contains 6,474 nodes and 13,895 edges with 100 timestamps in total.

\textbf{Stochastic Block Model (SBM) dataset\footnote{\url{https://github.com/IBM/EvolveGCN/tree/master/data}}}: It is generated using the Stochastic Block Model (SBM) model. The first snapshot of the dynamic graph is generated to have three equal-sized communities with in-block probability 0.2 and cross-block probability 0.01. To generate subsequent graphs, it randomly picks 10-20 nodes at each timestep and move them to another community. The final generated synthetic SBM graph contains 1000 nodes, 4,870,863 edges and 50 timestamps.

\textbf{Bitcoin-OTC (Bit-OTC) dataset\footnote{\url{http://snap.stanford.edu/data/soc-sign-bitcoin-otc.html}}}: It is who-trusts-whom network of people who trade using Bitcoin on a platform called Bitcoin OTC. The Bit-OTC dataset contains 5,881 nodes and 35,588 edges across 137 timestamps (weighted directed graph). This dataset exhibits highly transient dynamics.

\textbf{UC Irvine messages (UCI) dataset\footnote{\url{http://konect.cc/networks/opsahl-ucsocial/}}}: It contains sent messages between the users of the online student community at the University of California, Irvine. The UCI dataset contains 1,899 nodes and 59,835 edges across 88 timestamps (directed graph). This dataset too exhibits highly transient dynamics.

\textbf{Slashdot dataset\footnote{\url{http://konect.cc/networks/slashdot-threads/}}}: It is a large-scale social reply network for the technology website Slashdot. Nodes represent users and edges correspond to the replies of users. The edges are directed and start from the responding user. Edges are annotated with the timestamp of the reply. The Slashdot dataset contains 50, 824 nodes and 42, 968 edges across 12 timestamps.

\textbf{Facebook dataset\footnote{\url{https://data.mendeley.com/datasets/4dwzvcdsv3/2}}}: It is a large-scale Facebook wall post dynamic network, where each node is a user and a temporal directed edge represents a post from one user on another user’s wall at a given timestamp. The entire temporal graph contains 46,873 nodes and 857,815 edges over 30 timestamps sampled monthly from 14 October 2004 to 21st January 2009. 

\textbf{Reality Mining dataset\footnote{\url{http://realitycommons.media.mit.edu/realitymining.html}}}: The network contains human contact data among 100 students of the Massachusetts Institute of Technology (MIT); the data was collected with 100 mobile phones over 9 months in 2004. Each node represents a student; an edge denotes the physical contact between two nodes. In our experiment, the dataset contains 96 nodes and 1,086,403 undirected edges across 90 timestamps.

\textbf{Digg dataset\footnote{\url{http://konect.cc/networks/munmun_digg_reply/}}}: It is a large-scale dynamic reply network of the social news website Digg. Nodes represent the users of the Digg website while edges are used to describe the replied message actions between users. It contains 87,626 nodes and 30,398 directed edges over 90 timestamps.

\subsection{Performance evaluation for temporal link prediction} 
\label{p_linkpredict}
\subsubsection{Evaluation metrics} To evaluate our DynG2G model and demonstrate its effectiveness, we adopt two standard metrics:
\textbf{a) Mean average precision (MAP)}, where we rank all the predictions of node $q$ in a decreasing order of the probabilities. Then, we calculate the average precision ($AP$) for each node with the formula shown in Eq.~\ref{eq:MAP_definitions}, where $m$ is the number of connections predicted by our model for a particular node $q$,  $n$ is the maximum number of connections to be considered, $P(k)$ denotes the precision at $k$-th node, $rel(k) = 1$ if $k$-th node is connected to node $q$, otherwise 0. $MAP$ can be computed by averaging the $AP$ values of all the nodes $Q$ (see Eq.~\ref{eq:MAP_definitions}). 
\begin{equation}
\centering
MAP = \frac{1}{Q}\sum_q^Q{AP(q)}; \,\, AP(q) =\frac{1}{m} \sum_{k=1}^n{[P(k)\times rel(k)]}.
\label{eq:MAP_definitions}
\end{equation}
\textbf{b) Mean reciprocal rank (MRR)}. To calculate the $MRR$, we rank all the predictions of node $q$ in decreasing order of the probabilities, then we calculate the reciprocal rank ($1/k_q$) of node $q$ after finding the rank position ($k_q$) of the first relevant linked node for the query node $q$. Finally, we take the average reciprocal rank of all nodes as the $MRR$ value; see the detailed formula in Eq.~\ref{eq:MRR_definitions}. Note that since $MRR$ only considers the single highest-ranked relevant node, $MRR$ can give a general measure of quality for link prediction task; however, $MAP$ considers whether all of the relevant nodes tend to get ranked highly.
\begin{equation}
\centering
\label{eq:MRR_definitions}
MRR = \frac{1}{Q}\sum_{q\in Q}{\frac{1}{k_q}}
\end{equation}
\subsubsection{Implementation details}
Our implementation is based on the PyTorch framework, and all experiments were conducted on the NVIDIA Quadro RTX 6000 GPU (two 2.4 GHz 32 Core Processors; 1024GB DDR4 3200MHz Memory). We applied the Adam algorithm with hyperparameters (learning rate = 1e-3, tolerance = 100, one hidden layer of size 512, and epochs = 700) to minimize the triplet-based square and exponential loss for different embedding size ($L$ = 16, 32, 64, 128, 256). For the link prediction, the MLP model contains only one hidden layer with the same size as the node embedding size ($L$), and the learning rate is 1e-4. 

With the aforementioned two metrics, we evaluate our DynG2G model using the temporal link prediction task on eight benchmarks described in Section~\ref{sec:dataset}. The main purpose of temporal link prediction is to predict the links at timestamp $T+1$ using the transformed graph sequence embeddings up to timestamp $T$. Specifically, we first utilize the DynG2G model to obtain node embeddings ($h_i^{t}$) with k-hop neighbor factor ($K=2$) for all timestamps of each benchmark. The timestamps in each benchmark were first split into train/validation/test by ratios of 70\%, 10\% and 20\% for link prediction; the specific timestamp numbers can be seen in Table~\ref{tab:benchmark_table}. Before training the MLP model for the link prediction, we performed ``positive link'' and ``negative link'' sampling similar to \cite{bojchevski2017deep} for each timestamp in the training dataset. After that, each link feature was constructed as a 1-D vector, i.e., concatenate its nodes' embedding vectors as a single vector. Similarly, the validation and test link datasets were also built by the same procedure. Finally, we trained a MLP model with weighted cross entropy loss for predicting the link probability in the test timestamps. We evaluate the temporal link prediction performance with different graph embedding sizes ($L \in [16, 32, 64, 128, 256]$).

We have carried out experiments for eight temporal benchmarks and performed comparisons with other baseline methods including DynGEM~\cite{goyal2018dyngem}, EvolveGCN~\cite{pareja2020evolvegcn}, dyngraph2vecAE and dyngraph2vecAERNN~\cite{goyal2020dyngraph2vec} when possible. We list the final results on temporal link predictions of MAP and MRR in  Table~\ref{tab:method_comparisons_LP}. 
The MAP values of DynG2G are based on the best means computed over five initializations and over time; the best means correspond to different embedding dimension $L$, which herein we call it the optimum embedding dimension $L_o$. For the MRR values, we report the values for that embedding dimension, which may not be necessarily the best values.
DynG2G shows substantially better performance in terms of both metrics (MAP and MRR) for temporal link prediction compared to all other baseline methods, especially for the highly dynamic Bit-OTC dataset and UCI dataset. We note that for the UCI benchmark, the highest value obtained with DynG2G is 
$0.0347 = 0.0209 + 0.0138$, which is much higher than the second best $0.0270$ obtained by EvolveGCN. For the last four benchmarks there are no published results in the literature so we could not make direct comparisons with respect to the accuracy.
However, we managed to run EvolveGCN for two benchmarks only as shown in Table~\ref{tab:method_comparisons_LP}.
\begin{table}[ht]
\caption{Comparison results of temporal link prediction task for eight different benchmarks. DynG2G MAP values correspond to the best mean values
(averaged over five intializations and over time) selected over all embedding dimensions. MRR values correspond to the same embedding dimension, which we call
optimum $L_o$ herein, see Table~\ref{tab:DvsL}.}
\label{tab:method_comparisons_LP}
\centering
\fontsize{6.6}{8}\selectfont
\begin{tabular}{llll}
\toprule
Benchmark              & Method                     & MAP             & MRR    \\
\midrule
\multirow{5}{*}{AS}  & DynGEM                     & 0.0529          & 0.1028 \\
                     & dyngraph2vecAE             & 0.0331          & 0.1028 \\
                     & dyngraph2vecAERNN          & 0.0711          & 0.0493 \\
                     & EvolveGCN                  & 0.1534          & 0.3632 \\
                     & \textbf{DynG2G}            & \textbf{0.3154}   & \textbf{0.3880}   \\
\midrule
\multirow{5}{*}{SBM} & DynGEM                     & 0.168           & 0.0139 \\
                     & dyngraph2vecAE             & 0.0983          & 0.0079 \\
                     & dyngraph2vecAERNN          & 0.1593          & 0.012  \\
                     & EvolveGCN                  & 0.1989          & 0.0138 \\
                     & \textbf{DynG2G}            & $\mathbf{0.5146\pm0.0435}$   & $\mathbf{0.0294 \pm 0.0021}$\\
\midrule
\multirow{5}{*}{Bit-OTC} & DynGEM                     & 0.0529          & 0.0921                     \\
                             & dyngraph2vecAE             & 0.0090          & 0.0916 \\
                             & dyngraph2vecAERNN & 0.0220 & 0.1268                     \\
                              & EvolveGCN                  & 0.0028          & 0.0968                     \\
                             & \textbf{DynG2G}                   & \textbf{0.0662}          &\textbf{0.4804}                          \\
\midrule
\multirow{5}{*}{UCI}         & DynGEM                     & 0.0209          & 0.1055                     \\
                             & dyngraph2vecAE             & 0.0044          & 0.0540                     \\
                             & dyngraph2vecAERNN          & 0.0205          & 0.0713                     \\
                             & EvolveGCN                  & 0.0270          & 0.1379                     \\
                             & \textbf{DynG2G}                     & $\mathbf{0.0209\pm 0.0138}$          & $\mathbf{0.1842\pm 0.0766}$                    \\
\midrule
\multirow{1}{*}{Slashdot}  & \textbf{DynG2G}                     & $\mathbf{0.1637\pm 0.0227}$         & $\mathbf{0.4219\pm 0.0059}$                     \\
\midrule
\multirow{1}{*}{Facebook}   & \textbf{DynG2G}                     & $\mathbf{0.1335\pm 0.0260}$          & $\mathbf{0.2967\pm 0.0216}$                     \\
\midrule
\multirow{2}{*}{RealityM}    & \textbf{DynG2G}  &  \textbf{0.0266}  & \textbf{0.0597}                     \\
                                & EvolveGCN                  & 0.0090          & 0.0416 \\
\midrule
\multirow{2}{*}{Digg}     & \textbf{DynG2G}     & $\mathbf{0.0024\pm 0.0004}$   & $\mathbf{0.2210 \pm 0.0380}$   \\                  
                            & EvolveGCN                  &  0.0000126  &  0.0017 \\ 
\bottomrule
\end{tabular}
\end{table}

\begin{figure*}[!t]
\centering
\includegraphics[width= .8\linewidth]{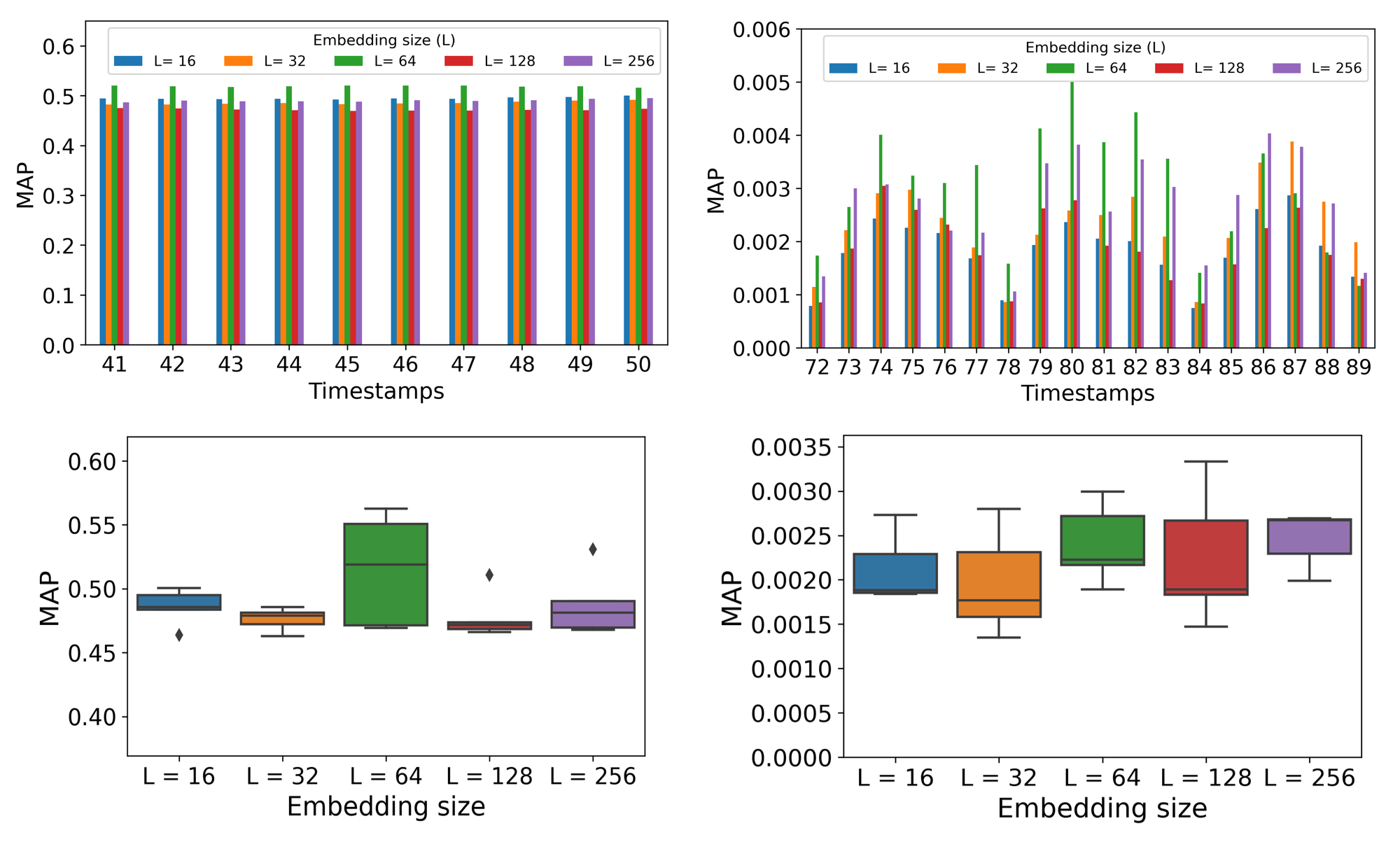}
\caption{MAP results for the temporal link prediction tasks with DynG2G (first row) on synthetic SBM dataset (left) and Digg dataset (right) for different embedding sizes ($L$ = 16, 32, 64, 128, 256). These results correspond to one specific initialization.  In the second row, we show the MAP statistical results over five runs and over all timestamps. Results for the three other benchmarks are shown in Appendix Fig. S2 while details of the five initialization runs are shown in Fig. S3. The SBM graph evolves smoothly across different snapshots, however, the Digg graph exhibits highly transitional dynamics across snapshots at different timestamps as shown in Fig. \ref{fig:dataset}.} 
\label{fig:lp_SBM_UCI_MAP}
\end{figure*}

These eight benchmarks have different temporal dynamics whose complexity varies with time. In order to gain some insight into their dynamics, we plot the MAP metric for different embedding size $L$ as a function of time for the testing period for one initialization for the SBM and Digg datasets in Fig.~\ref{fig:lp_SBM_UCI_MAP}; many more detailed results are shown in the Appendix (Fig. S3), where we carry out five runs for each benchmark
corresponding to different initializations.
As we see, the value of MAP depends on the embedding dimension $L$ and it is rather stable as a function of time for the SBM benchmark.
However, for the highly transient Digg data set the MAP metric varies significantly in time, and different embedding sizes yield the best MAP value depending on the specific timestamp. The best embedding size, $L$, may vary for different initializations (see Fig. S3), so in Fig.~\ref{fig:lp_SBM_UCI_MAP} in the bottom row we present the MAP statistical results over five runs and over all timestamps. From these plots, we see that the best $L=64$ for both the SBM and the Digg benchmarks, which is not possible to infer from a single initialization. 
%
Overall this behavior is consistent with the insights provided by the uncertainty quantification analysis as we explain in the next section. 

\subsection{Uncertainty quantification for node embeddings}
The existing temporal graph embedding methods are \textit{deterministic} and they transform the temporal graph nodes as fixed points in the latent space, hence ignoring important information associated with node uncertainty. However, DynG2G is based on Graph-2-Gauss (G2G) static graph embedding method~(\cite{bojchevski2017deep}) and can provide uncertainty quantification, which is important for downstream analytics tasks. Specifically, G2G learns graph embeddings as multivariate Gaussian distributions including the mean (``position'') and variance (``uncertainty'') vectors assigned to each node; the variance plays a key role in measuring the uncertainty of graph node representation. Moreover, as shown in \cite{bojchevski2017deep} and also in \cite{xu2020understanding}, DynG2G can track the stable and unstable dimensions in time and hence obtain the effective dimensionality of the system, which is related to the required embedding size $L$. For temporally evolving graphs, this dimensionality may be changing in time, but DynG2G can capture in a quantified manner the transient dynamics of the evolving system. Hence, it is not surprising that the different datasets in the eight benchmarks exhibit different accuracy for different embedding sizes $L$ as a function of time (e.g., Digg, Bit-OTC and UCI benchmarks) whereas the benchmarks with smoother transient dynamics exhibit a more stable pattern in time with respect to $L$. 
\begin{figure*}[!t]
\centering
\includegraphics[width = .9\textwidth]{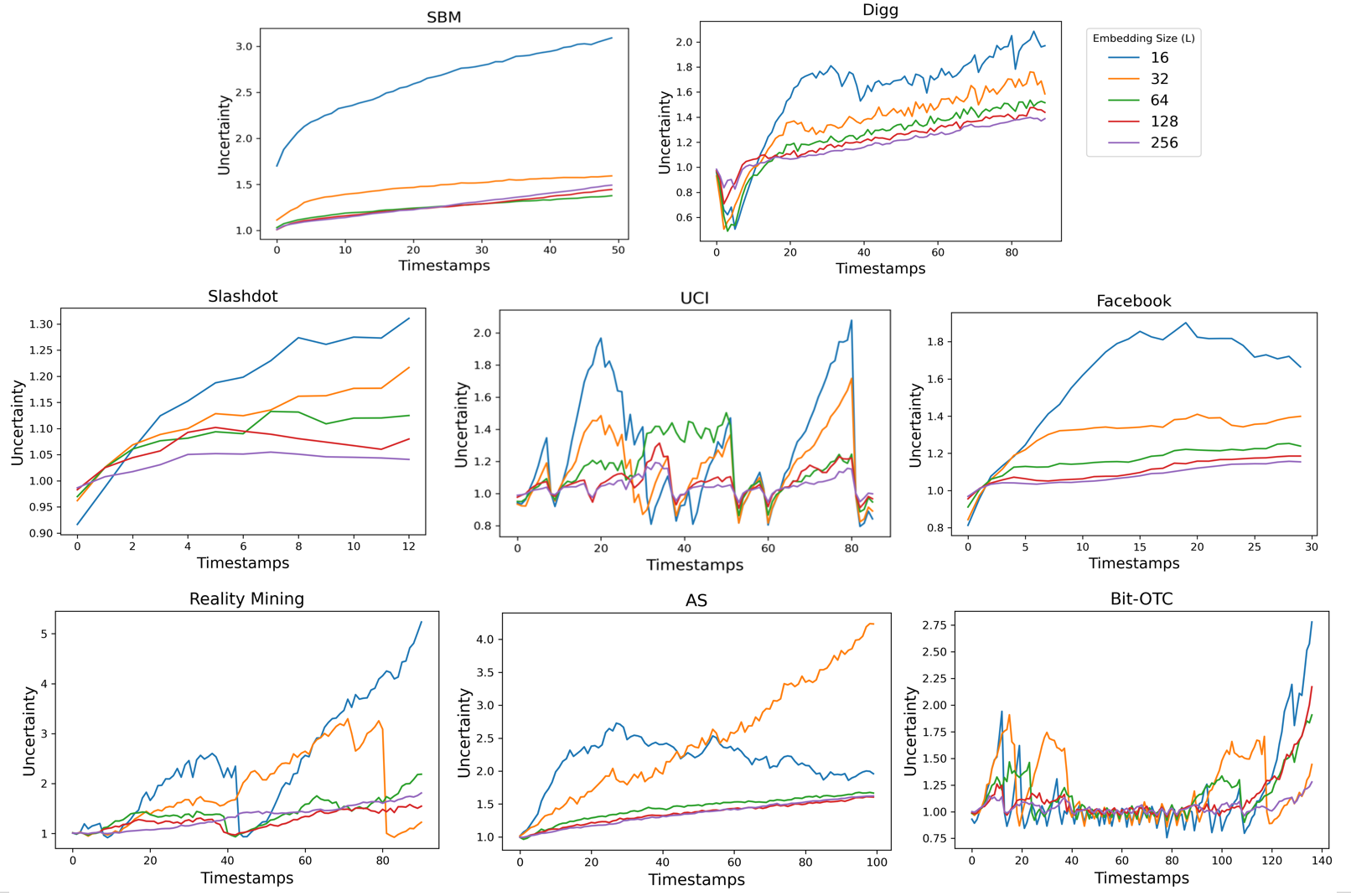}
\caption{Uncertainty versus time for temporal node embeddings with DynG2G for the eight benchmarks with embedding size ($L$ = 16, 32, 64, 128, 256). The X-axis represents the timestamps for each benchmark, while the Y-axis refers to the averaged standard deviations computed by the variances predicted by DynG2G for all nodes at different timestamps. These results correspond to a single initialization but more detailed results for all five initializations are shown in the Appendix Fig. S4.
}
\label{fig:uncertainty_8benchmarks}
\end{figure*}

Correspondingly, the total variance as a function of time, plotted in Fig.~\ref{fig:uncertainty_8benchmarks}, shows a similar dependence on the embedding size $L$. Interestingly, for the SBM benchmark there is a clear separation of the values of variance for $L < 64$ but good convergence with respect to $L$ for $L > 64$, and therefore the effective dimensionality of the SBM system is $L=64$, which does not change in time. The smooth and slow variation of uncertainty versus time for SBM is consistent with the stable results on MAP versus time shown in Fig.~\ref{fig:lp_SBM_UCI_MAP}(a). We have observed similar trends for the AS benchmark, and here too the effective embedding size is $L=64$ and does not change in time, consistent with the trends of the variance (uncertainty) versus time in Fig. ~\ref{fig:uncertainty_8benchmarks}. In contrast for the UCI and Bit-OTC benchmarks, we see that the variance exhibits a different pattern and seems to converge for $L\ge128$ for the UCI and for $L=256$ for the OTC. In fact, for the latter the variance increases exponentially at longer times. This, in turn, reveals that both the UCI and Bit-OTC are high-dimensional systems with highly transient dynamics so it is not surprising that the MAP accuracy varies also in time as a function of different embedding sizes. Finally, we note that the results presented in Fig.~\ref{fig:uncertainty_8benchmarks} are only for one initialization and a more systematic study is needed, as we present in the Appendix for different initializations (Fig. S4), to infer the proper value of the effective
uncertainty dimensionality, which we denote as $D_u$, corresponding to the smallest uncertainty. It is interesting to note that the curves of uncertainty versus time look similar for different initialization runs as shown in the Appendix (Fig. S4), unlike the MAP versus times curves for different values of $L$.

\begin{table}[ht]
\caption{Correlation between the optimal embedding size ($L_o$) and the effective dimensionality of uncertainty ($D_u$) for all eight benchmarks. This correlation is based on averaging the results over all timestamps and over five different runs corresponding to different initializations. These results suggest a clear path to selecting the graph optimum embedding dimension by choosing $L_o \ge D_u$.}
\label{tab:DvsL}
\centering
\begin{tabular}{llll}
\toprule
Benchmark  & Optimal embedding size & UQ dimensionality \\
            &  ($L_o$)               &  ($D_u$)         \\
\midrule
AS             & 64        & 64                    \\
SBM            & 64        & 64                    \\
UCI            & 256       & 256                   \\
Bit-OTC        & 256       & 256                   \\
Slashdot       & 64        & 64                    \\
Facebook       & 32        & 32                    \\
Reality Mining & 64        & 64                    \\
Digg           & 64        & 64                     \\
\bottomrule
\end{tabular}
\end{table}



\section{Conclusion}
We proposed an efficient stochastic dynamic graph embedding method (DynG2G), which enables us to automatically learn dynamic graph representations in lower-dimensional ``function'' space with high efficiency. The learned graph embedding can be used for diverse downstream graph learning tasks, e.g., recommendation, fraud detection,
and search. 
The current methods mostly rely on large autoencoder models (DynGEM, dyngraph2vecAE) or memory based architectures that use RNN/LSTM for memory in addition to encoder-decoder networks (dyngraph2vecAERNN). EvolveGCN makes use of Graph Convolutional networks (GCN) along with GRU and LSTM, and it consists of stacked layers of convolutional networks. DynG2G, in comparison, uses a single layer neural network and at different timesteps of the graph, and we use the learned weights from the previous timestep, which enables the model to converge faster by transferring temporal information.

We computed all eight benchmarks in about 0.6 to 60 seconds per epoch on a single GPU, see Table \ref{tab:DynG2G_Time} (Appendix) for detailed computational costs of different embedding sizes; 
see also Fig. \ref{fig:S6} for the GPU memory footprint for two of the benchmarks. Moreover, we computed the computational cost for the benchmark ``Reality Mining" using EvolveGCN and we found that it takes about $3.40$ seconds per epoch versus $0.698$ seconds for DynG2G for embedding size $L=16$. 
We obtained experimental results based on a range of graphs from 96 to 87626 nodes with both slowly varying and rapidly changing dynamics in order to test DynG2G for diverse applications with corresponding graphs characterized by multi-rate temporal dynamics. For example, we focused on the SBM and Digg benchmarks as the former has only 1,000 nodes but close to 4.9 million edges while the latter has 87,626 nodes and only 30,398 edges. Moreover, SBM exhibits stable dynamics whereas Digg exhibits rapidly evolving dynamics. The other six benchmarks shown in the Appendix have characteristics that fall between these two benchmarks. We have demonstrated that our DynG2G model outperforms other baseline methods (for available benchmark metrics) and effectively learns highly informative graph embeddings for even graphs with rapidly changing dynamics (i.e., Digg, Facebook, Reality Mining, Bit-OTC and UCI benchmarks). 

The novelty of our approach is that DynG2G in addition to the mean values it can simultaneously predict the node embedding uncertainty, which plays a crucial role in quantifying the evolving uncertainty and complexity of dynamical systems over time. In particular, we obtained a ``universal" relation of the optimal embedding dimension, $L_o$, versus the effective dimensionality of uncertainty, $D_u$, and we found that $L_0=D_u$ for all cases. To the best of our knowledge, this is the first time that such a relation is obtained for graph embedding of temporal graphs of arbitrary size, and it reveals that the uncertainty quantification approach we employ in the DynG2G algorithm correctly captures the {\em intrinsic dimensionality} of the dynamics of such evolving graphs despite the diverse nature and composition of the graphs at each timestamp.

A current limitation of DynG2G is that it employs a fixed embedding dimension, which is pre-defined by the user. However, based on our finding that a different embedding
size is optimal at different timestamps, a new adaptive strategy combined with self-attention mechanism can be designed in future work to allow {\em variable embedding size} as the temporal dynamics of the systems evolves, especially for systems like the Digg, UCI and Bit-OTC benchmarks. Using the newly discovered $L_o - D_u$ correlation, we can develop a new criterion on how to select the optimum graph
embedding dimension $L_o$ on-the-fly by quantifying the effective dimensionality of corresponding node uncertainty $D_u$. 
We expect that this will lead to increasing both accuracy and efficiency of DynG2G. Another limitation of the model that is still very challenging to deal with is the missing timestamp data problem for real world datasets, e.g., electronic healthcare record database. To this end, DynG2G has to be combined with 
state-of-the-art data imputation techniques in future work. Moreover, adopting a loss function such as the one from TransE \cite{bordes2013translating} could extend our DynG2G to heterogeneous graphs.
Finally, it will be interesting to develop a similar method but based on hyperbolic embeddings and examine if the dimensionality reduction is even greater.

\appendix
\renewcommand{\thefigure}{S\arabic{figure}}
\renewcommand{\thetable}{S\arabic{table}}
\setcounter{figure}{0}
\setcounter{table}{0}
\section{}
We present additional results for all benchmarks to supplement the results in the main text in Figs. S1-S4. We present results for the MRR metric in Fig. S5. In addition, we present the specific epochs-wise computational cost (in seconds) in Table~\ref{tab:DynG2G_Time} corresponding to different embedding sizes ($L = 16, 32, 64, 128, 256$) for DynG2G over 5 runs.
\begin{figure*}[!t]
\centering
\includegraphics[width=.95\linewidth]{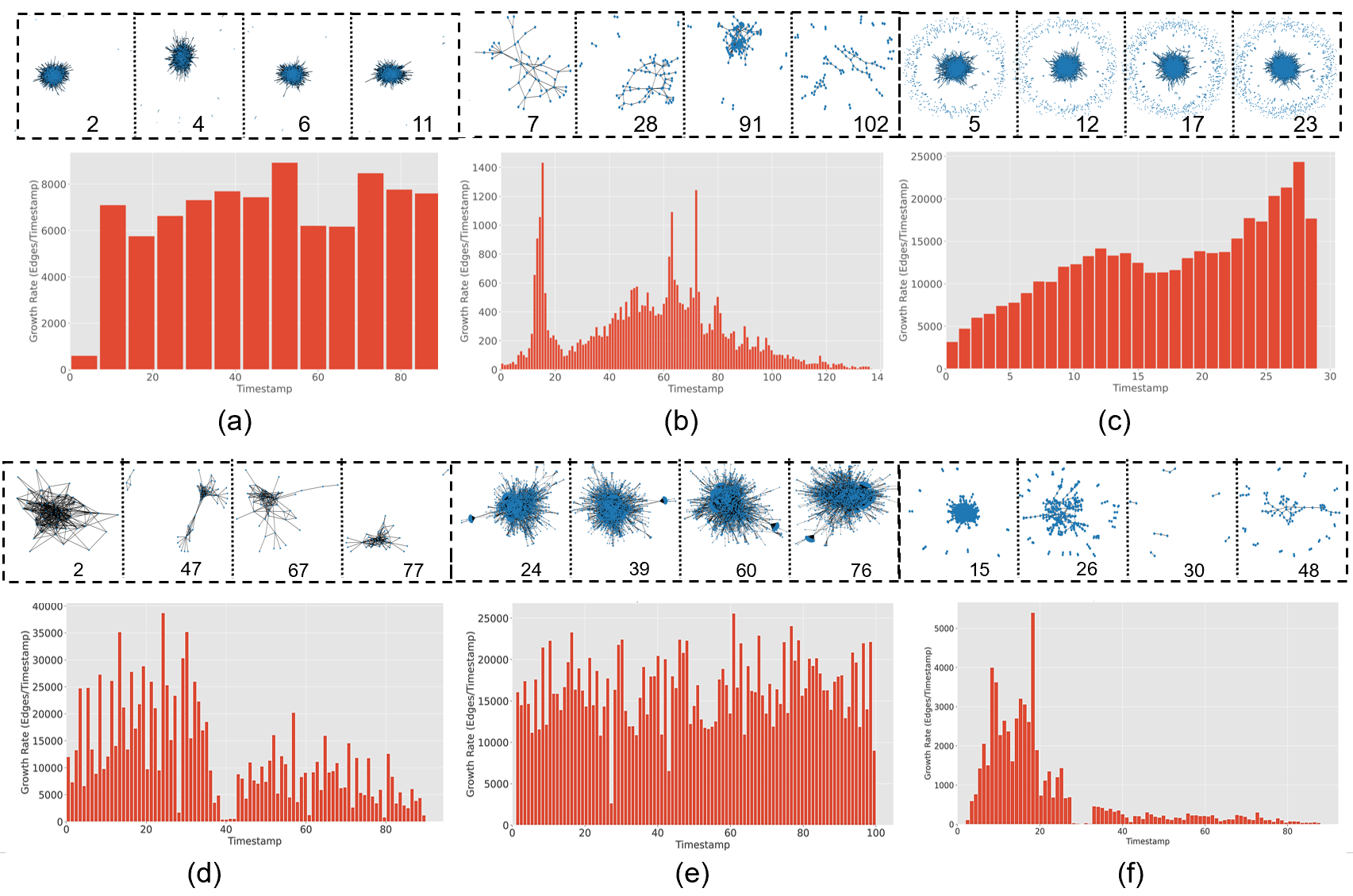}
\caption{Visualizations of the evolving dynamics of the remaining six temporal graphs in our benchmarks. (a) Slashdot, (b) Bit-OTC, (c) Facebook, (d) Reality Mining, (e) AS, (f) UCI. In each sub-figure, the top row shows four random snapshots of the temporal graph at different timestamps (indicated by the number in each image); the bottom row shows the growth rate of edges as a function of time in each benchmark. (a), (c) and (e) exhibit stable dynamics whereas (b), (d) and (f) exhibit rapidly changing dynamics.}
\label{fig:S1}
\end{figure*}

\begin{figure*}[!t]
\centering
\includegraphics[width=.95\linewidth]{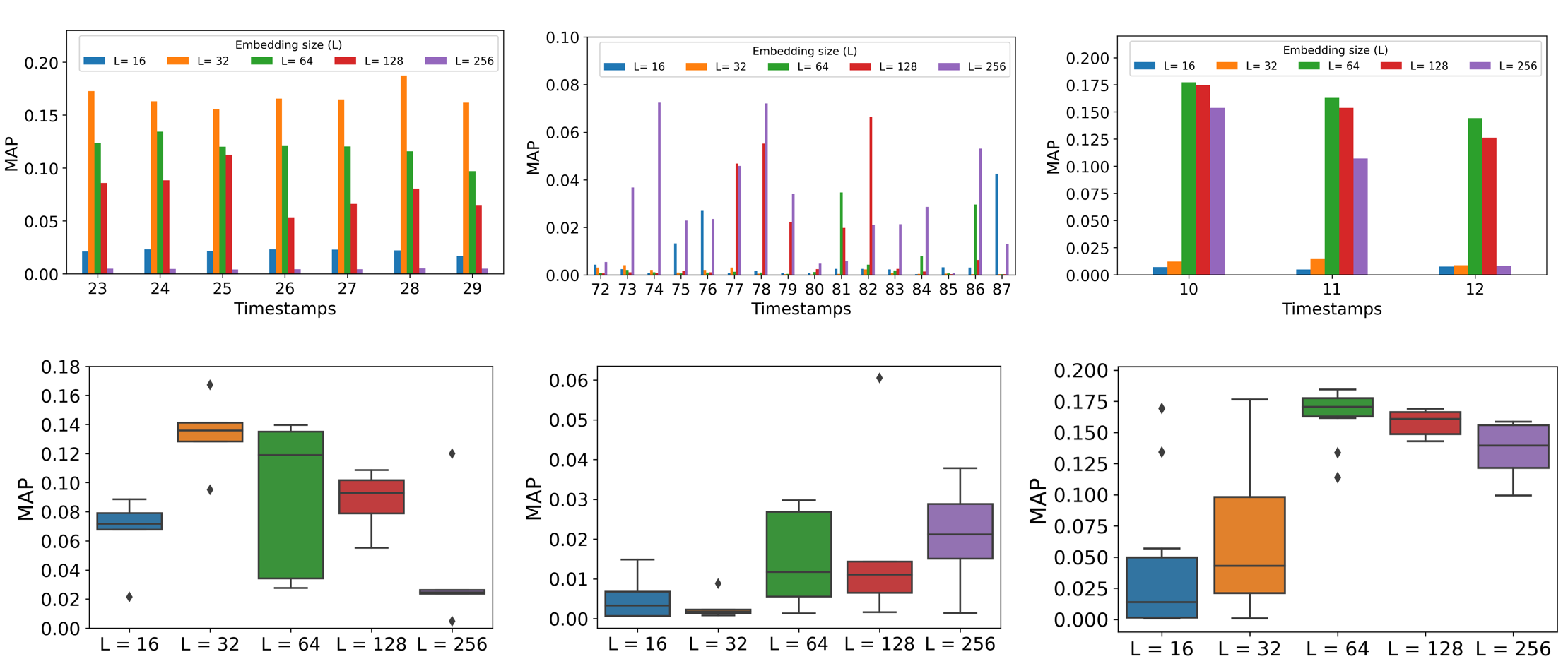}
\caption{MAP results for the temporal link prediction tasks with DynG2G for three other benchmarks (Facebook-\textit{left column}, UCI-\textit{middle column }and Slashdot-\textit{right column}) with different embedding size ($L$ = 16, 32, 64, 128, 256). In the first row we show the MAP results for the link prediction versus different timestamps for one initialization. In the second row, we show the MAP statistical results over five runs (corresponding to different initializations) and over all timestamps for Facebook, UCI and Slashdot datasets.
}
\label{fig:S2}
\end{figure*}

\begin{figure*}[!t]
\centering
\includegraphics[width=\linewidth]{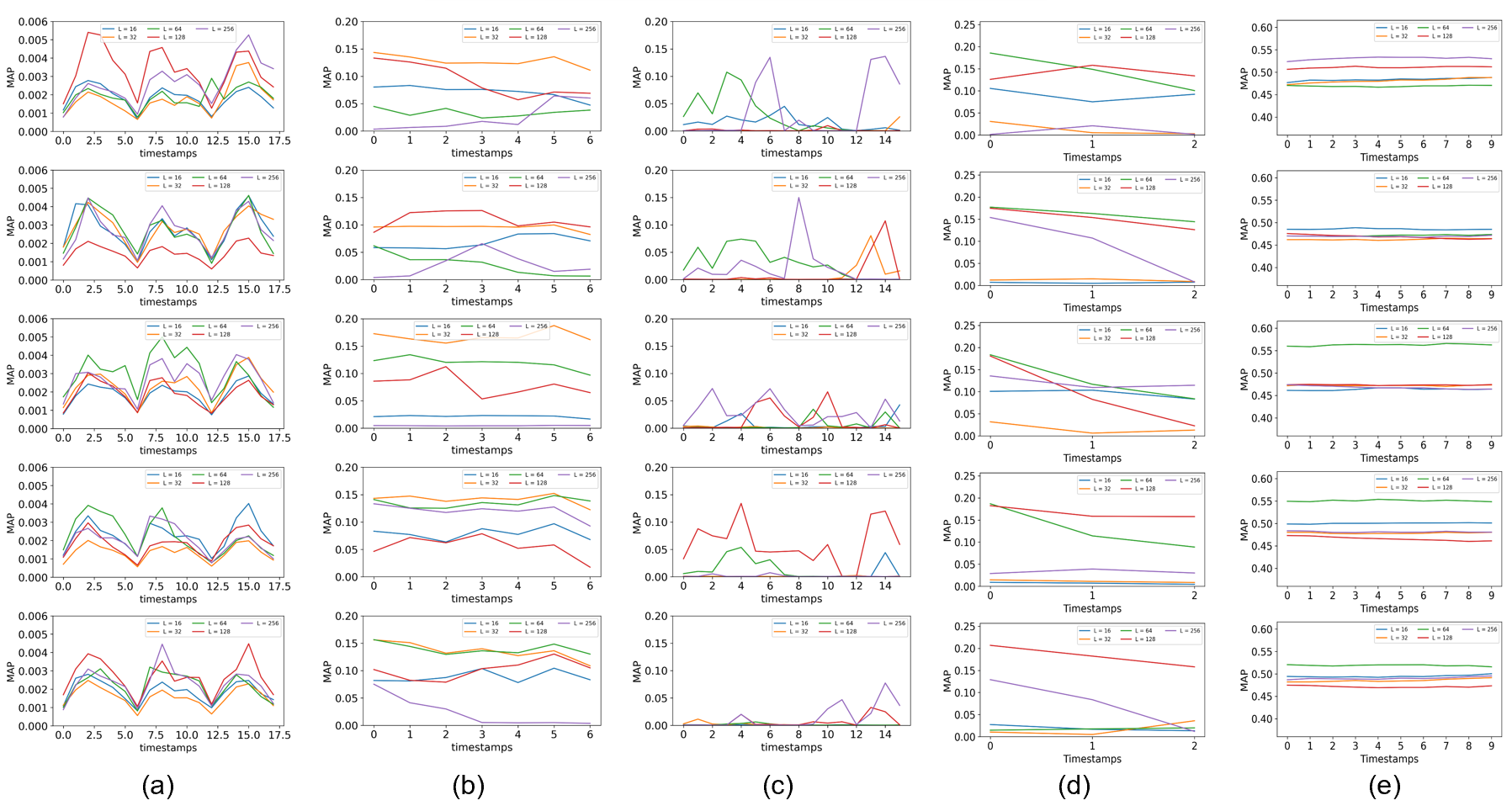}
\caption{MAP results using DynG2G method for temporal link prediction tasks for five different benchmarks with different embedding sizes ($L = 16, 32, 64, 128, 256$). {Columns (a)-(e) refer to the Digg, Facebook, UCI, Slashdot and SBM benchmark, respectively}. For each benchmark, we performed the temporal link prediction task with 5 different random initializations, as shown in the five different rows.}
\label{fig:S3}
\end{figure*}

\begin{figure*}[!t]
\centering
\includegraphics[width=\linewidth]{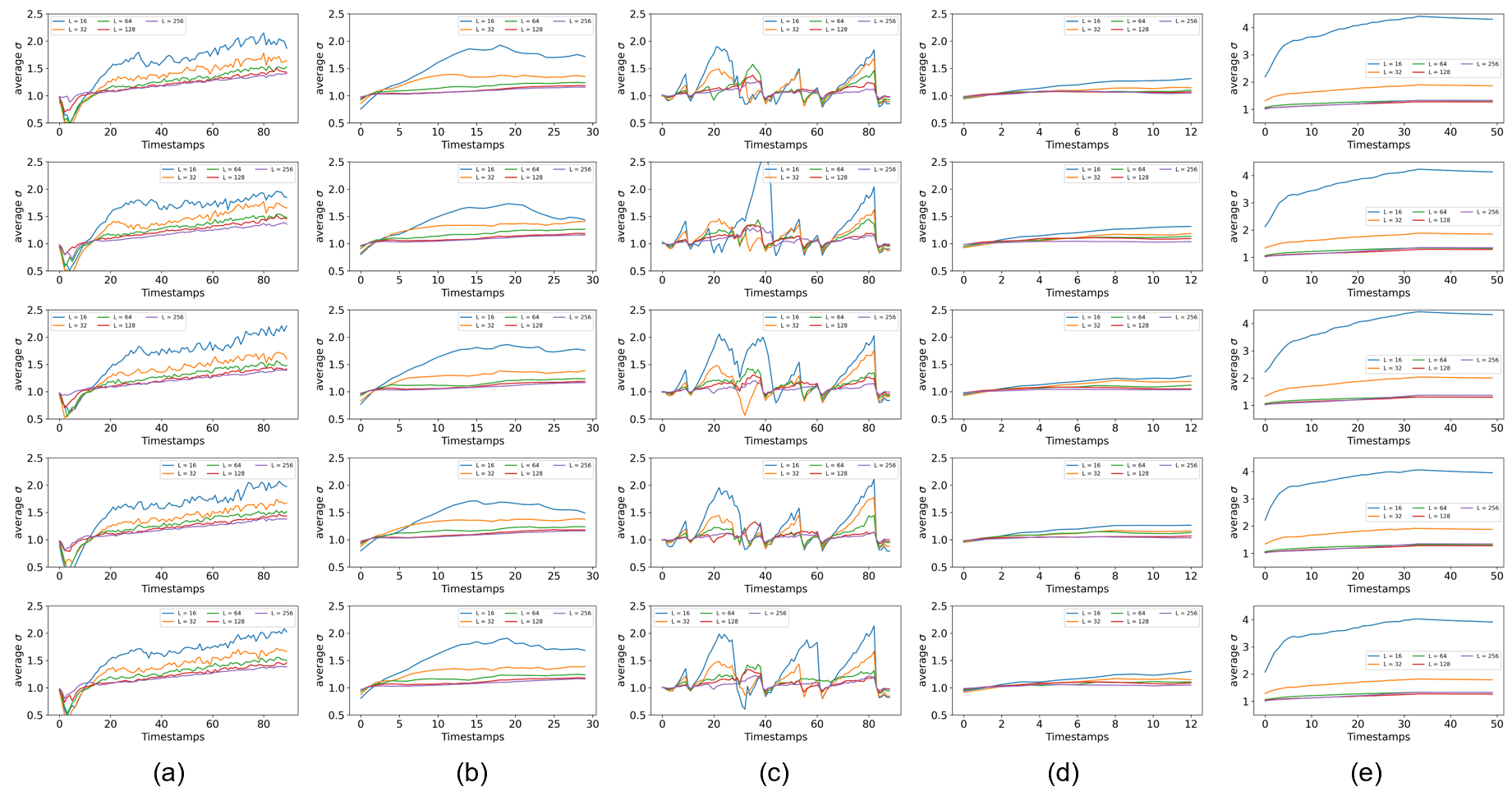}
\caption{Uncertainty (standard deviation) versus time for 5 runs (corresponding to different initializations) 
 for temporal link prediction tasks with DynG2G for five different benchmarks with different embedding sizes ($L = 16, 32, 64, 128, 256$). {Columns (a)-(e) refer to the Digg, Facebook, UCI, Slashdot and SBM benchmark, respectively}. For each benchmark, we performed the temporal link prediction task with 5 different random initializations, as shown in the five different rows. The uncertainty is independent of the initialization and the variation is relatively smooth for each $L$ except the UCI benchmark that exhibits highly irregular dynamics.}
\label{fig:S4}
\end{figure*}

\begin{figure*}[!t]
\centering
\includegraphics[width=\linewidth]{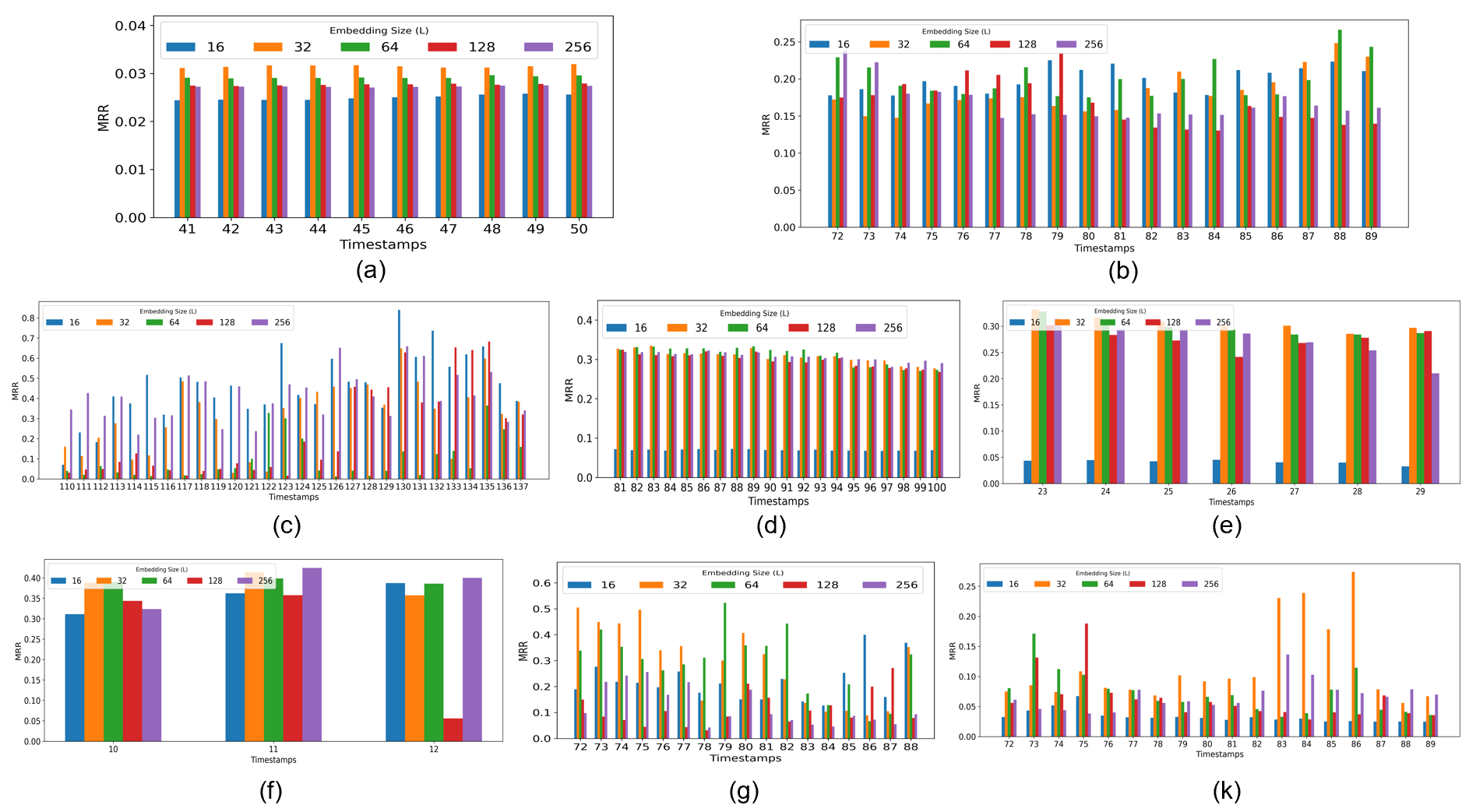}
\caption{{MRR results using DynG2G method for temporal link prediction tasks on eight different benchmarks. (a) SBM, (b) Digg, (c) Bit-OTC, (d) AS, (e) Facebook, (f) Slashdot, (g) UCI, and (k) Reality Mining.}
}
\label{fig:S5}
\end{figure*}
\begin{table*}[h!]
\caption{Average and standard deviation of the epoch-wise computational cost (in seconds) corresponding to different embedding sizes ($L = 16, 32, 64, 128, 256$) for DynG2G over 5 runs on the NVIDIA Quadro RTX 6000 GPU (two 2.4 GHz 32 Core Processors and 1024GB DDR4 3200MHz Memory).}
\label{tab:DynG2G_Time}
\centering
\begin{tabular}{llllll}
\toprule
Benchmark  & L = 16 & L = 32 & L = 64 & L = 128 & L = 256 \\
\midrule
SBM             & $\mathbf{9.363 \pm 0.524}$  & $\mathbf{9.480 \pm  0.402}$  
                & $\mathbf{9.228 \pm 0.495}$  & $\mathbf{9.218 \pm 0.334}$
                & $\mathbf{8.960 \pm 0.245}$
                \\
UCI             & $\mathbf{1.104 \pm 0.010 }$ & $\mathbf{1.076 \pm 0.014}$  
                & $\mathbf{1.071 \pm 0.011}$ & $\mathbf{1.075 \pm 0.010}$ 
                & $\mathbf{1.086 \pm 0.010}$
                \\
Bit-OTC         & $\mathbf{3.533 \pm 0.094}$ & $\mathbf{3.871 \pm 0.063}$  
                & $\mathbf{3.818 \pm 0.251}$ & $\mathbf{3.904 \pm 0.717}$
                & $\mathbf{3.806 \pm 0.159}$ 
                \\
Slashdot        & $\mathbf{33.915 \pm 0.133}$ & $\mathbf{33.768 \pm 0.171}$ 
                & $\mathbf{33.932 \pm 0.232}$ & $\mathbf{34.203 \pm 0.252}$ 
                & $\mathbf{34.573 \pm 0.281}$
                \\
Facebook        & $\mathbf{58.543 \pm 10.321}$ & $\mathbf{58.687 \pm 10.272}$
                &$\mathbf{ 59.112 \pm 10.251}$ & $\mathbf{59.517 \pm 10.253}$
                & $\mathbf{60.452 \pm 10.271}$ 
                \\
Reality Mining  & $\mathbf{0.698 \pm 0.069}$ & $\mathbf{0.668 \pm 0.071}$ 
                & $\mathbf{0.670 \pm 0.071}$ & $\mathbf{0.672 \pm 0.071}$
                & $\mathbf{0.675 \pm 0.071}$ 
                \\
Digg            & $\mathbf{23.465 \pm 0.189}$ & $\mathbf{23.504 \pm 0.233}$
                & $\mathbf{23.632 \pm 0.339}$ & $\mathbf{24.230 \pm 1.190}$
                & $\mathbf{26.232 \pm 3.268}$
                \\
\bottomrule
\end{tabular}
\end{table*}

\begin{figure*}[h!]
\centering
\includegraphics[width=.8\linewidth]{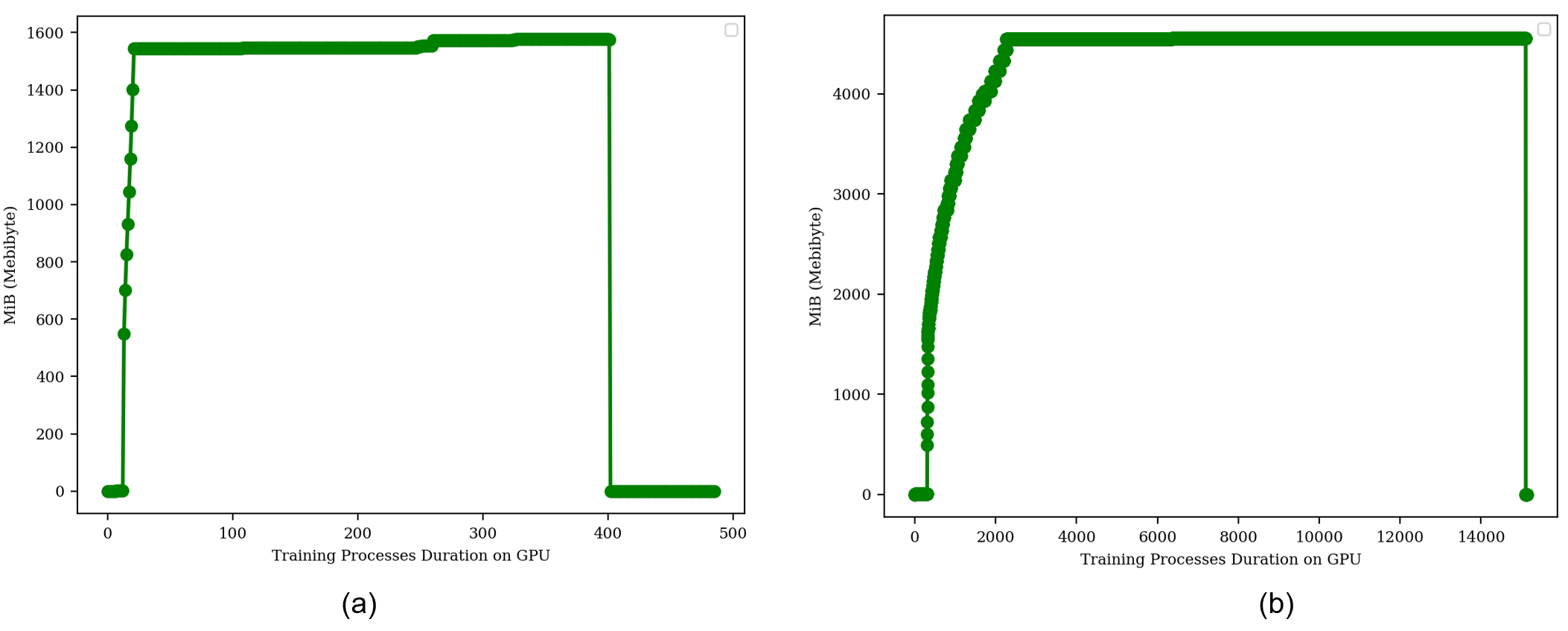}
\caption{GPU Memory footprint plots for the proposed DynG2G model trained with two different sizes of dynamic graph benchmarks (Digg and Reality Mining). (a) and (b) show the amount of main memory that DynG2G code uses for learning graph embeddings for the Reality Mining dataset (a) and Digg dataset (b). Reality Ming uses 1.6 GB Maximum while Digg uses ~5 GB Maximum but it is persistent usage.}
\label{fig:S6}
\end{figure*}


\end{document}